\documentclass[a4paper, 10pt, conference]{IEEEtran}

\IEEEoverridecommandlockouts

\usepackage{amsmath}
\usepackage[pdftex]{graphicx}
\graphicspath{{figures/}}
\usepackage{hyperref}
\hypersetup{
colorlinks=true,
linkcolor=black,
citecolor=black,
filecolor=black,
urlcolor=black,
}
\usepackage[T1]{fontenc}
\usepackage[utf8]{inputenc}
\usepackage{csquotes}
\usepackage[english]{babel}
\usepackage[font=small]{caption}
\usepackage[font=small]{subcaption}
\usepackage{siunitx}
\usepackage{upgreek}

\newcommand{\norm}[1]{\left\lVert#1\right\rVert}

\title{
Self-Supervised Flow Estimation using Geometric Regularization with Applications to Camera Image and Grid Map Sequences
\thanks{\IEEEauthorrefmark{1}Both authors contributed equally.}
}

\author{
\IEEEauthorblockN{Sascha Wirges\IEEEauthorrefmark{1} \& Qiuhao Zhang}
\IEEEauthorblockA{
Mobile Perception Systems Group\\
FZI Research Center for Information Technology\\
Karlsruhe, Germany\\
wirges@fzi.de
}
\and
\IEEEauthorblockN{Johannes Gräter\IEEEauthorrefmark{1} \& Christoph Stiller}
\IEEEauthorblockA{
Institute of Measurement and Control Systems\\
Karlsruhe Institute of Technology (KIT)\\
Karlsruhe, Germany\\
\{johannes.graeter,stiller\}@kit.edu}
}

\begin{document}

\maketitle
\thispagestyle{empty}
\pagestyle{empty}

\begin{abstract}
We present a self-supervised approach to estimate flow in camera image and top-view grid map sequences using fully convolutional neural networks in the domain of automated driving.
We extend existing approaches for self-supervised optical flow estimation by adding a regularizer expressing motion consistency assuming a static environment.
However, as this assumption is violated for other moving traffic participants we also estimate a mask to scale this regularization.
Adding a regularization towards motion consistency improves convergence and flow estimation accuracy.
Furthermore, we scale the errors due to spatial flow inconsistency by a mask that we derive from the motion mask.
This improves accuracy in regions where the flow drastically changes due to a better separation between static and dynamic environment.
We apply our approach to optical flow estimation from camera image sequences, validate on odometry estimation and suggest a method to iteratively increase optical flow estimation accuracy using the generated motion masks.
Finally, we provide quantitative and qualitative results based on the KITTI odometry and tracking benchmark for scene flow estimation based on grid map sequences.
We show that we can improve accuracy and convergence when applying motion and spatial consistency regularization.
\end{abstract}

\section{Introduction} \label{sec:introduction}

\begin{figure}[t]
\centering
\def\svgwidth{\columnwidth}
\begin{footnotesize}
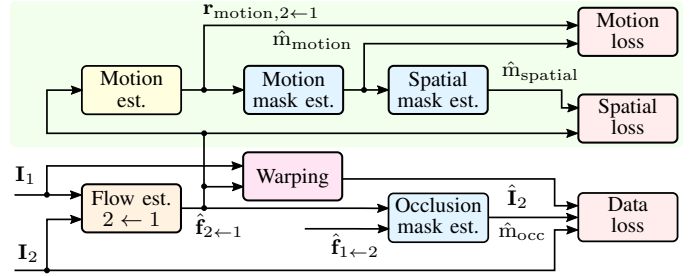
\end{footnotesize}
\caption{
Overview of our training strategy for self-supervised flow estimation.
Our goal is to estimate the flow $\hat{\mathbf{f}}_{2 \leftarrow 1}$ from two measurements $\mathbf{I}_1$ and $\mathbf{I}_2$ that transforms coordinates from frame 1 into frame 2.
During training, we determine the data loss from the warped image $\hat{\mathbf{I}}_2$ and $\mathbf{I}_2$, scaled by an occlusion mask.
Our main contribution is to add a motion inconsistency loss which is based on the residual of $\hat{\mathbf{f}}_{2 \leftarrow 1}$ and a robustly estimated motion flow, scaled by a learned motion mask $\hat{\mathrm{m}}_{\mathrm{motion}}$ (green area).
Compared to previous work \cite{Meister2018} we also improve the spatial inconsistency loss by deriving a mask $\hat{\mathrm{m}}_{\mathrm{spatial}}$ from $\hat{\mathrm{m}}_{\mathrm{motion}}$.
Note that the backward direction $2 \leftarrow 1$ and mask regularizers are omitted here for simplicity.
}
\label{fig:overview}
\end{figure}

% General intro
An accurate reconstruction of the environment is an essential component of mobile robotic systems such as automated vehicles.
To achieve this, measurements from heterogenous sensors such as cameras, lidars or radars are fused into one common representation which provides redundancy in case of sensor outage or occlusions.
Part of this reconstruction is the precise motion estimation of relevant traffic participants.
Whereas radars can directly measure radial velocities, motion in subsequent camera and lidar measurements needs to be inferred from corresponding features.

% Definition: Scene & Optical Flow
This process is known as \textit{scene flow estimation}.
Similar to Vedula et al.~\cite{Vedula2005} scene flow may be defined as the three-dimensional movement of features in subsequent measurements due to motion.
Projecting scene flow into a camera yields \textit{optical flow} which may also be interpreted as the movement of image features between subsequent camera images.

% Applications
In this work we aim to estimate flow for subsequent camera images and top-view multi-layer grid maps.
The latter can be interpreted as an orthogonal projection of 3D features onto the ground surface.
This means that images from cameras and grid maps only differ in the projection model being used.
Then, given two subsequent multi-channel images, our system outputs an optical flow and an odometry estimate which describes the vehicle's motion over time.

% Supervised Deep Learning methods
In recent years, accurate and efficient flow estimators based on convolutional neural networks emerged (see \cite{Dosovitskiy2015, Ilg2017, Sun2018}).
However, most of these methods need labeled training data, either simulated or annotated by humans.
Whereas the realism of simulated data is still suboptimal, manual annotation of training data is cumbersome and tedious.

% What we do here
Therefore, we present our approach to learn convolutional flow estimators without the need of manually annotated training data.
It is based on \textit{Unflow} \cite{Meister2018}, a self-supervised learning framework for optical flow which we extend with objectives to assure motion and spatial consistency.

% Content Summary
After a discussion of related work in Section~\ref{sec:related_work} we present our objectives which enable self-supervised training in Section~\ref{sec:objectives} and derive comprehensive loss functions that are used in the training stage.
We then introduce two applications of our flow estimation method where we implement the objectives introduced and show experimental results on the KITTI odometry benchmark.
Section~\ref{sec:application_to_camera_image_sequences} introduces our approach for optical flow estimation in camera images.
We then present the flow estimation based on grid map sequences in Section~\ref{sec:application_to_grid_map_sequences} where we conduct a quantitative evaluation based on the KITTI odometry and tracking benchmark.
Finally, we conclude our work and propose ideas for future research in Section~\ref{sec:conclusion}.

\section{Related Work} \label{sec:related_work}
Scene flow and odometry estimation are closely related.
First, we describe scene flow estimation methods which establish feature correspondences.
Then, we provide an overview on odometry estimation methods which operate based on these correspondences.

\subsection{Scene Flow Estimation} \label{sec:related_work_optical_and_scene_flow_estimation}
% Sparse Optical Flow and ICP
Optical flow estimation is fundamental for all camera-based motion estimation algorithms.
It has been studied extensively since the beginning of computer vision and is still topic of recent research (see \cite{Lucas1981, Horn1981, Rublee2011}).

% Dense Optical Flow
Whereas early work mostly focuses on sparse optical flow estimation, dense optical flow estimation has become popular with the work of Farneback et al.~\cite{Farneback2003}.
Traditionally, dense methods suffer from high computational cost and low estimation accuracy in low-structured image regions.
A major breakthrough was achieved by applying convolutional neural networks (CNNs) to this problem, rendering flow estimation feasible for complete images with increased accuracy.
Especially the work of Dosovitskiy et al.~\cite{Dosovitskiy2015} and the follow up of Ilg et al.~\cite{Ilg2017} show impressive results by introducing FlowNet, a stacked encoder-decoder architecture where certain parts of the network account for large displacement and others for details. 
Sun et al.~\cite{Sun2018} adopt important parts of their work and present a pyramid-based architecture that produces competitive results while drastically reducing memory.
Whereas Ilg et al. use supervised training, Sun et al. exploit the representation of the problem as motion of pixels between images.
Instead of annotating the optical flow, images are transformed to one common frame using the estimated optical flow such that the photometric error is minimized.
In that way, optical flow can be estimated without the need of annotating data, only using camera images.
This methodology of unsupervised optical flow learning is also used by Meister et al.~\cite{Meister2018} to train the architecture proposed by Ilg et al. without using annotated data.

\subsection{Odometry Estimation} \label{sec:related_work_odometry_estimation}
Estimating flow solves the problem of finding correspondences between subsequent measurements.
This information can be used to estimate odometry.
Given subsequent measurements, odometry estimation is the task of determining transformations $\mathbf{T}(t)$ over time due to motion.
Therefore, a set of corresponding features between subsequent measurements is used to determine transformations which minimize the distance between these features.
Additional constraints, e.g. the optimization on a manifold of the complete motion space, enable high robustness and accuracy of the odometry estimation.

\paragraph{Based on Rigid Body Kinematics}
Given a set of 2D / 3D correspondences, Horn et al. \cite{Horn1988} present a closed-from solution to estimate transformations $\boldsymbol{T} \in SE(3)$ which minimize the sum of squared distances.
The translational component of $\boldsymbol{T}$ can be determined by estimating the average translation of all correspondences.
To estimate the rotational component, the authors suggest an orthonormal matrix decomposition, maximizing the similarity between transformed correspondences.
An analytical and robust approach using a weighted Least-Squares was presented by Marx \cite{Marx2017} to mitigate the influence of false correspondences on the final result.

\paragraph{Based on Epipolar Geometry}
Having obtained a set of 2D correspondences of the scene, the ego motion can be estimated up to scale.
Therefore, the motion estimate lies on a manifold with five degrees of freedom~\cite{Graeter2017} which are the three angles of rotation and the two translation vector directions.
Classically, an error metric that is based on the epipolar geometry is used for optimization (see \cite{Hartley2003, Nister2004}).
In regions where the image gradient is perpendicular to the optical flow, optical flow estimation becomes problematic since the optical flow can not be observed.
This is called the aperture problem~\cite{Hartley2003}.
The motion, however, supplies additional information about the the scene which reduces this effect.
Consequently, recent contributions aim to solve motion and optical flow estimation jointly, approaching from different directions.
Bradler et al.~\cite{Bradler2017} integrate an epipolar geometry based motion constraint into the optimization problem of the optical flow estimation and solve it jointly.
Engel et al. (\cite{Engel2014, Engel2018}) circumvent the explicit optical flow estimation by representing the optical flow as an unscaled three dimensional reconstruction and the motion.
The problem is solved by minimizing the photometric error by techniques known from Simultaneous Localization and Mapping (SLAM). 
This idea was adapted by Zhou et al.~\cite{Zhou2017} and in the follow up of Casser et al.~\cite{Casser2018} that propose an architecture to estimate the three dimensional structure and the motion jointly with CNNs.

\section{Objectives} \label{sec:objectives}
Following up on the work of Meister et al.~\cite{Meister2018} we revise the objectives for self-supervised training of flow estimators and present two novel loss terms that assure motion and spatial consistency. 
Given two subsequent frames 1 and 2, the estimated image coordinate
\begin{equation}
\hat{\boldsymbol{x}}_2 = \hat{\mathbf{f}}_{2 \leftarrow 1}(\boldsymbol{x}_1)
\end{equation}
in frame 2 corresponds to the image coordinate in frame 1 transformed by the corresponding flow.
Here, $\hat{\mathbf{f}}_{2 \leftarrow 1}$ depicts the estimated flow from frame 1 to frame 2.
In the following, we create coordinate-wise loss terms
\begin{align}
\mathcal{L}_{2 \leftarrow 1}(\boldsymbol{x}_1) &= \mathcal{L}_{\mathrm{data}, 2 \leftarrow 1}(\boldsymbol{x}_1) \nonumber \\
&+ \mathcal{L}_{\mathrm{motion}, 2 \leftarrow 1}(\boldsymbol{x}_1) \nonumber \\
&+ \mathcal{L}_{\mathrm{spatial}, 2 \leftarrow 1}(\boldsymbol{x}_1)
\end{align}
to penalize data-, motion- and spatial inconsistency which we present below.
Similarly we add these loss terms in temporal backward direction (frame 2 to frame 1) such that the final loss
\begin{equation}
\mathcal{L} = \sum_{\boldsymbol{x}_1 \in \mathcal{I}_1} \mathcal{L}_{2 \leftarrow 1}(\boldsymbol{x}_1) + \sum_{\boldsymbol{x}_2 \in \mathcal{I}_2} \mathcal{L}_{1 \leftarrow 2}(\boldsymbol{x}_2)
\end{equation}
penalizes data-, motion- and spatial inconsistency in temporal forward and backward direction based on sets of valid coordinates $\mathcal{I}_1$ and $\mathcal{I}_2$.

\subsection{Data Consistency} \label{sec:objectives_data_consistency}
We denote the pixel values in frame 1 and frame 2 as $\mathbf{p}_1(\boldsymbol{x})$ and $\mathbf{p}_2(\boldsymbol{x})$, respectively.
By claiming data consistency between two corresponding pixels, we assume that the residual
\begin{equation}
\mathbf{r}_{\mathrm{data}, 2 \leftarrow 1}(\boldsymbol{x}_1) = \mathbf{p}_1(\boldsymbol{x}_1) - \mathbf{p}_2(\hat{\boldsymbol{x}}_2),
\end{equation}
which models the difference between the pixel in frame 1 and the pixel at the transformed coordinate in frame 2, is small.

This assumption, however, is violated in the presence of occlusions.
Therefore, we estimate an occlusion mask $\hat{\mathrm{m}}_{\mathrm{data}}$ as proposed by Meister et al. \cite{Meister2018}.
Then, we define the coordinate-wise loss function
\begin{align}
\mathcal{L}_{\mathrm{data}, 2 \leftarrow 1}(\boldsymbol{x}_1) &= \hat{\mathrm{m}}_{\mathrm{data}, 1}(\boldsymbol{x}_1)~\uprho\left(\norm{\mathbf{r}_{\mathrm{data}, 2 \leftarrow 1}(\boldsymbol{x}_1)}^2\right) \nonumber \\
&+ \upepsilon\left(\hat{\mathrm{m}}_{\mathrm{data}, 1}(\boldsymbol{x}_1)\right)
\end{align}
where we chose the robustifier
\begin{equation}
\uprho(x) = \left(x + 10^{-3}\right)^{0.45}
\end{equation}
to be the generalized charbonnier loss~\cite{Sun2014}.
Here and analogously in the following sections we add a regularizer
\begin{equation}
\upepsilon(x) = \uprho\left(\left(1 - x\right)^2\right)
\end{equation}
to avoid trivial solutions, e.g. where $\hat{\mathrm{m}}_{\mathrm{data}}(\boldsymbol{x}) = 0$ for all $\boldsymbol{x} \in \mathcal{I}$.

\subsection{Motion Consistency} \label{sec:objectives_motion_consistency}
As the sensor moves, the flow corresponding to the static environment is consistent to the sensor motion.
Hence, we define a motion model $\mathbf{f}_{\mathrm{motion}, 2 \leftarrow 1}$ to map image coordinates from frame 1 to frame 2 to model the sensor motion.
The implementation of $\mathbf{f}_{\mathrm{motion}}$ and its estimation depends on the projection model used and is presented in Sections~\ref{sec:application_to_camera_image_sequences_implementation} based on the epipolar geometry in camera images and in Section~\ref{sec:application_to_grid_map_sequences_implementation} based on the rigid body motion in top-view grid maps.
Given the motion model, we assume that the residual
\begin{equation} \label{eq:objectives_motion_residual}
\mathbf{r}_{\mathrm{motion}, 2 \leftarrow 1}(\boldsymbol{x}_1) = \hat{\mathbf{f}}_{\mathrm{motion}, 2 \leftarrow 1}(\boldsymbol{x}_1) - \hat{\mathbf{f}}_{2 \leftarrow 1}(\boldsymbol{x}_1),
\end{equation}
modeling the difference between motion and scene flow, is small for static environment.
However, this assumption is violated for dynamically changing environment, e.g. other traffic participants.
Therefore, we estimate a motion mask $\hat{\mathrm{m}}_{\mathrm{motion}}$ similar to Zhou et al.~\cite{Zhou2017} that scales down the residuals representing non-static environment.
The final motion loss
\begin{align}
\mathcal{L}_{\mathrm{motion}, 2 \leftarrow 1}(\boldsymbol{x}_1) &= \hat{\mathrm{m}}_{\mathrm{motion}, 1}(\boldsymbol{x}_1)~\uprho\left(\norm{\mathbf{r}_{\mathrm{motion}, 2 \leftarrow 1}(\boldsymbol{x}_1)}^2\right) \nonumber \\
&+ \upepsilon\left(\hat{\mathrm{m}}_{\mathrm{motion}, 1}(\boldsymbol{x}_1)\right)
\end{align}
then penalizes motion that is not consistent with the flow of the static environment.

\subsection{Spatial Consistency} \label{sec:objectives_spatial_consistency}
A basic assumption in this work is that the environment consists of rigid objects.
Depending on the projection model, the optical flow of linearly moving rigid objects may be similar.
With this in mind, we assume that the spatial flow gradient
\begin{equation}
\mathbf{r}_{\mathrm{spatial}, 2 \leftarrow 1}(\boldsymbol{x}_1) = \boldsymbol{\nabla}\hat{\mathbf{f}}_{2 \leftarrow 1}(\boldsymbol{x}_1)
\end{equation}
is small for rigidly connected objects.
Here, we use a second-order approximation for $\boldsymbol{\nabla}\hat{\mathbf{f}}(\boldsymbol{x})$.
However, at object boundaries the flow might change drastically.
To model this effect we determine a spatial mask $\hat{\mathrm{m}}_{\mathrm{spatial}}$ that scales the residuals.
Thus, the loss
\begin{align}
\mathcal{L}_{\mathrm{spatial}, 2 \leftarrow 1}(\boldsymbol{x}_1) &= \hat{\mathrm{m}}_{\mathrm{spatial}, 1}(\boldsymbol{x}_1)~\uprho\left(\norm{\mathbf{r}_{\mathrm{spatial}, 2 \leftarrow 1}(\boldsymbol{x}_1)}^2\right) \nonumber \\
&+ \upepsilon\left(\hat{\mathrm{m}}_{\mathrm{spatial}, 1}(\boldsymbol{x}_1)\right)
\end{align}
assures spatial consistency.
\section{Application to Camera Image Sequences} \label{sec:application_to_camera_image_sequences}
\begin{figure*}[ht]
\includegraphics[width=\linewidth]{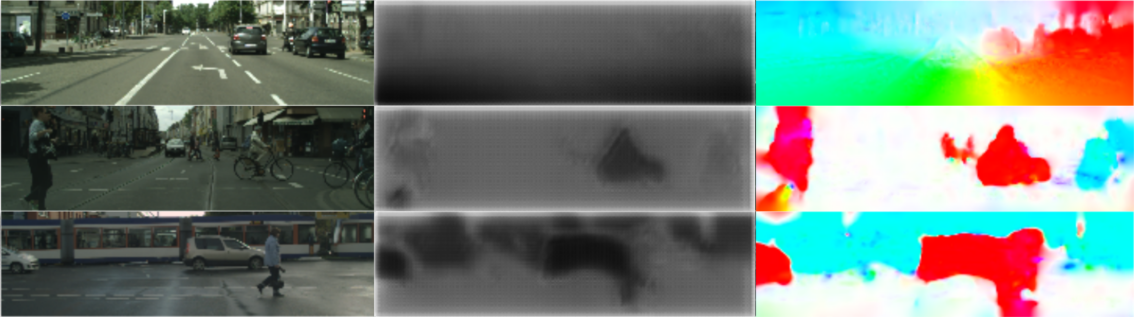}
\caption{
Results on the Cityscapes dataset for optical flow estimation on camera image sequences.
From left to right: Input camera images, estimated motion masks and optical flow estimates.
}
\label{fig:camera_masks}
\end{figure*}

\subsection{Implementation} \label{sec:application_to_camera_image_sequences_implementation}
We apply the joint estimation of optical flow and motion to consecutive camera images from a monocular camera.
To the frontend network which estimates the optical flow, we add a backend to estimate motion model and mask.
To extract motion, a separate encoder is used and its output is fed into a fully-connected layer, that performs regression on the five parameters roll, pitch and yaw angle of the camera as well as pitch and yaw angle of the translation vector.
The translation vector has length one, since the scale is difficult to observe as mentioned by Casser et al.~\cite{Casser2018}.
The motion consistency constraint described in section~\ref{sec:objectives_motion_consistency} is used as an additional error function to the data loss.
As explained in section~\ref{sec:objectives}, this constraint is only valid for pixels that correspond to the same moving structure and the influence of optical flow that corresponds to moving objects is reduced by $\hat{\mathrm{m}}_{\mathrm{motion}}$.
A crucial part is the regularization of this motion consistency mask so that it does not assign zero weight to all pixels.
Whereas Zhou et al.~\cite{Zhou2017} propose to punish differences to inlier probability one for each pixel, we calculate the epipolar error with the 8 point algorithm~\cite{Hartley2003} for each pixel and apply a soft threshold onto it.
By punishing the difference of this reference mask to the inlier probabilities, zero weight will be avoided.
To supply information which pixels are inliers to the network, the reference mask is stacked to the input of the encoder.
This scheme enables the extraction of moving objects: The inverse of the reference mask is stacked to the encoder input and the motion is estimated a second time.
In that manner object motion can be extracted and the part that does not belong to the moving object will be extracted by the inlier probabilities.
This scheme may be applied in a recursive manner for any number of objects, using the outliers of the previous step as input for the next step.

\subsection{Experiments} \label{sec:application_to_camera_image_sequences_experiments}
We validate our method on camera image sequences on the Cityscapes dataset~\cite{Cordts2016} and the KITTI dataset~\cite{Geiger2012}.
Whereas Cityscapes serves as pretraining data, sequences 00-07 of the KITTI odometry data set are used for finetuning and sequence 09 for validation.
We initialize the frontend weights with the pretrained weights supplied by Meister et al.~\cite{Meister2018}.
The backend is then trained for 10 epochs with batch size 4 and a learning rate of 5$\cdot\mathrm{10}^{\mathrm{-7}}$ with fixed frontend parameters.
Finally, the complete network is trained for 5 epochs.
Qualitative results are shown in Figure~\ref{fig:camera_masks}.
Figure~\ref{fig:results_odometry} depicts the estimated odometry based on the motion masks and the five-point-algorithm~\cite{Nister2004}.

\section{Application to Grid Map Sequences} \label{sec:application_to_grid_map_sequences}

Here, we represent the scene by a multi-layer top-view grid map which provides a mapping from two-dimensional discretized ground surface coordinates to higher-dimensional features.
Originally introduced by Elfes et al. \cite{Elfes1989}, grid maps are well-suited for sensor fusion and enable the use of efficient convolutional operations due to their dense grid structure.
We interpret grid maps as multi-channel images and use them as input for our flow estimation method.
As we use lidar measurements, we compose features out of the number of surface reflections, minimum and maximum height above ground and average reflection intensity.
By casting rays we also incorporate the height of shadows above ground from the sensor origin to the reflection positions.
More details on the grid mapping process are provided by Wirges et al.~\cite{Wirges2018}.

\begin{figure*}[t]
\centering
\begin{subfigure}[h]{0.19\linewidth}
\fbox{\includegraphics[width=0.93\linewidth]{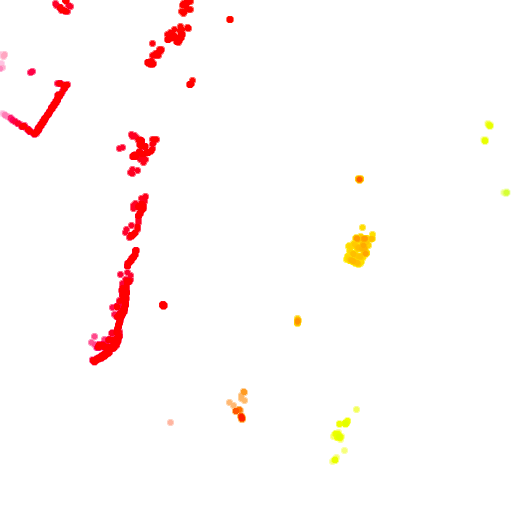}}
\caption{Scene flow}
\label{fig:results_scene_estimation}
\end{subfigure}
\begin{subfigure}[h]{0.19\linewidth}
\fbox{\includegraphics[width=0.93\linewidth]{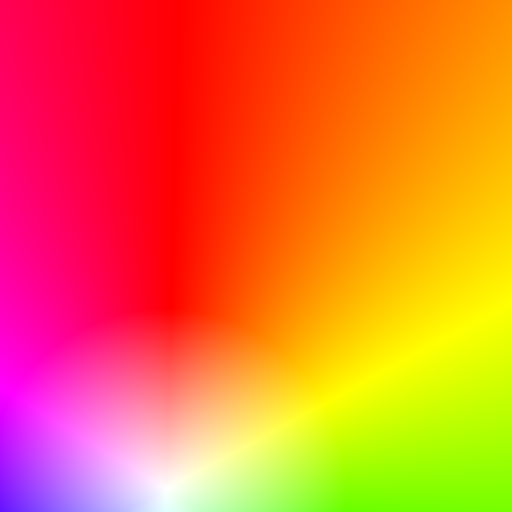}}
\caption{Motion flow}
\label{fig:results_motion_flow}
\end{subfigure}
\begin{subfigure}[h]{0.19\linewidth}
\fbox{\includegraphics[width=0.93\linewidth]{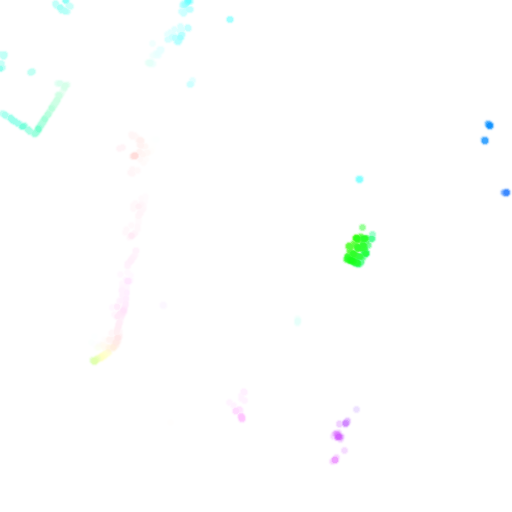}}
\caption{Compensated flow}
\label{fig:results_motion_compensated_flow}
\end{subfigure}
\begin{subfigure}[h]{0.19\linewidth}
\fbox{\includegraphics[width=0.93\linewidth]{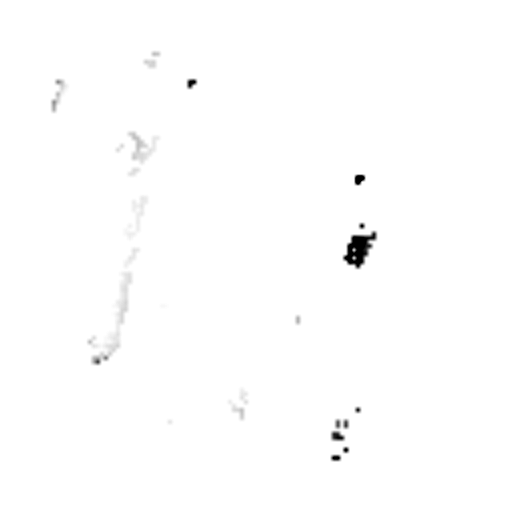}}
\caption{Motion mask}
\label{fig:results_motion_mask}
\end{subfigure}
\begin{subfigure}[h]{0.19\linewidth}
\fbox{\includegraphics[width=0.93\linewidth]{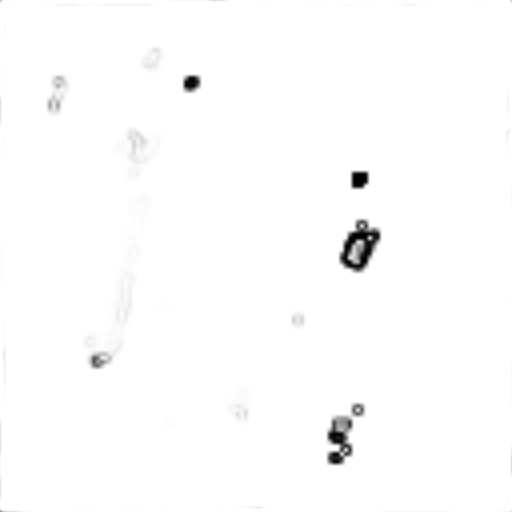}}
\caption{Spatial mask}
\label{fig:results_spatial_mask}
\end{subfigure}
\caption{
Explanation of the scene flow estimation process based on grid map sequences on an exemplary scenario.
Here, the vehicle is turning right where another vehicle ahead is driving in the same direction.
Fig.~\ref{fig:results_scene_estimation} describes the scene flow as it is estimated by our network.
Based on this scene flow, a rigid-body transformation is estimated in the training process that we transform into a motion flow field (see Fig.~\ref{fig:results_motion_flow}).
The difference of scene and motion flow is depicted in Fig.~\ref{fig:results_motion_compensated_flow}.
We observe that usually only moving obstacles show a large remaining flow.
From this compensated flow a mask is estimated (see Fig.~\ref{fig:results_motion_mask}) which we use to scale the data loss.
Fig.~\ref{fig:results_spatial_mask} depicts the mask to scale the spatial loss which is the gradient magnitude of the motion mask.
Best viewed digitally with zoom.
Figures~\ref{fig:results_scene_estimation}, \ref{fig:results_motion_flow} and \ref{fig:results_motion_compensated_flow} encode flow direction as hue and norm as value in HSV color space.
}
\label{fig:qualitative_results}
\end{figure*}

The scene flow estimation process is depicted in Figure~\ref{fig:qualitative_results}.
Given two subsequent frames of preprocessed multi-layer grid maps our system should estimate a transform describing the rigid-body motion between the two frames and a velocity for each grid cell in which reflections were detected.

\subsection{Implementation} \label{sec:application_to_grid_map_sequences_implementation}
In the following we implement our models based on the FlowNetC architecture presented in \cite{Dosovitskiy2015}.
Alternatively, we evaluated PWCNet as presented by Sun et al. \cite{Sun2018}.
However, we could not observe large benefits compared to FlowNetC in either accuracy or inference time due to computationally expensive correlations at each scale.

\paragraph{Preprocessing} \label{sec:application_to_grid_map_sequences_implementation_preprocessing}
As lidar reflections are sparse in the grid map, computing the data consistency loss based on inaccurate flow estimates may lead to vanishing gradient when warped reflections of one frame are not close to reflections in the other frame.
To mitigate this problem we apply Gaussian filtering to each grid map layer with a variance that is adapted during the training, leading to better convergence due to the non-vanishing gradient.

\paragraph{Receptive Field} \label{sec:application_to_grid_map_sequences_implementation_receptive_field}
In the KITTI odometry benchmark all inter-frame translations and rotations are below 2.5m and $5^{\circ}$, respectively.
With a grid map size of 60m$\times$60m and a cell size of 0.15m we then determine the receptive field size as 135 cells in width and height to fully cover these transformations.
Knowing the required receptive field size we remove the last two layers of the FlowNetC network and reduce the size by 40 \% compared to the original architecture.

\paragraph{Transformation Estimation} \label{sec:application_to_grid_map_sequences_implementation_transformation_estimation}
To reliably estimate a transformation between two frames, we use the weighted least squares method presented by Marx \cite{Marx2017} which was introduced in Section~\ref{sec:related_work_odometry_estimation}.
This method yields an orthonormal rotation matrix $\hat{\boldsymbol{R}}_{2 \leftarrow 1}$ and a translation $\hat{\boldsymbol{t}}_{2 \leftarrow 1}$ so that we can estimate the motion flow
\begin{equation}
\hat{\mathbf{f}}_{\mathrm{motion}, 2 \leftarrow 1}(\boldsymbol{x}_1) = \hat{\boldsymbol{R}}_{2 \leftarrow 1} \boldsymbol{x}_1 + \hat{\boldsymbol{t}}_{2 \leftarrow 1}
\end{equation}
as defined in Section~\ref{sec:objectives_motion_consistency}.

\paragraph{Mask Estimation} \label{sec:application_to_grid_map_sequences_implementation_mask_estimation}
We determine the motion mask
\begin{equation}
\hat{\mathrm{m}}_{\mathrm{motion}, 1}(\boldsymbol{x}_1) = 1 - \tanh\left(\norm{\mathbf{r}_{\mathrm{motion}, 2 \leftarrow 1}(\boldsymbol{x}_1)}^2\right)
\end{equation}
by applying the hyperbolic tangent function to the squared motion residual norm (see Equation~\ref{eq:objectives_motion_residual}).
We then compute the spatial mask as the normalized Sobel derivatives of the motion mask in both directions.

\subsection{Experiments} \label{sec:application_to_grid_map_sequences_experiments}

\begin{figure}[ht]
\centering
\def\svgwidth{0.95\columnwidth}
\begin{small}
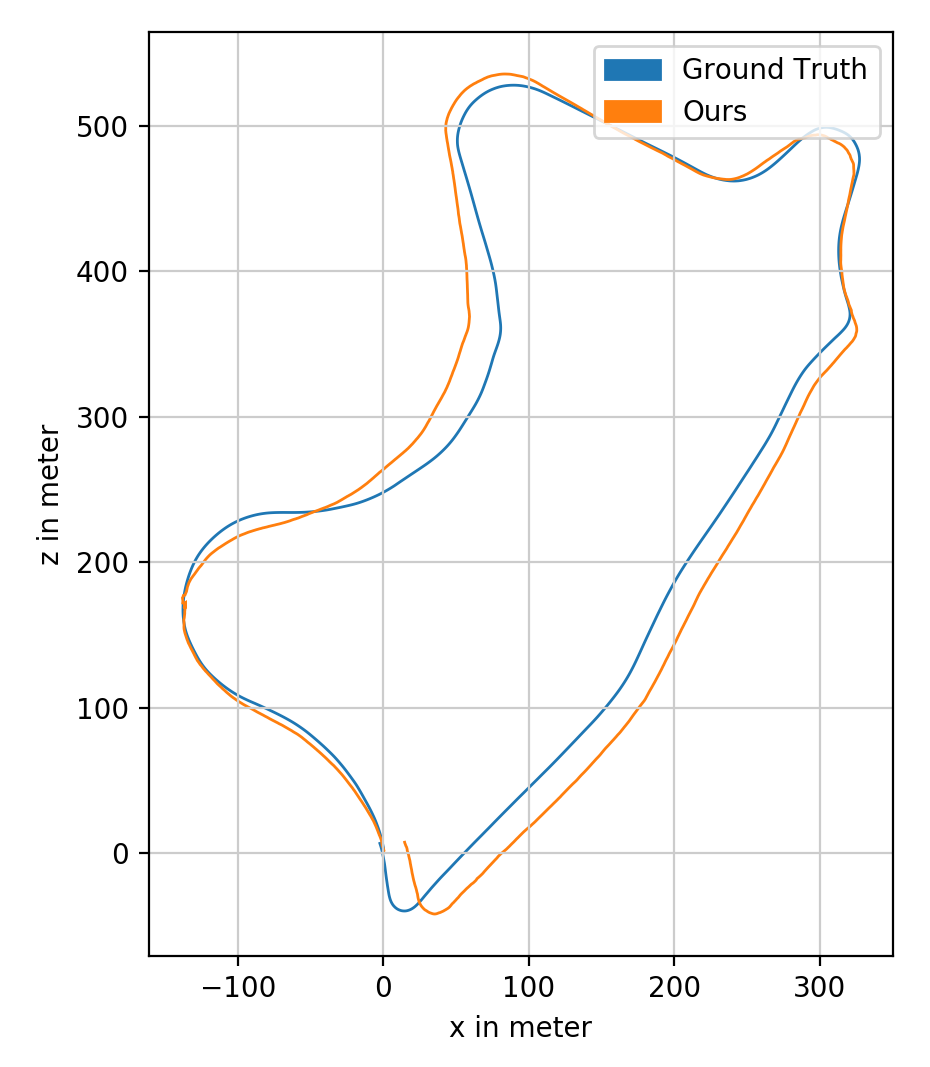
\end{small}
\caption{
Comparison of reference path (red) with odometry from grid map (green) and image sequences (blue) estimated by the motion model implementation for KITTI odometry sequence 9.
The scale for image sequences is obtained from the reference.
}
\label{fig:results_odometry}
\end{figure}

\sisetup{table-number-alignment=center, exponent-product=\cdot, output-decimal-marker = {.}}
\begin{table}[ht]
\centering
\begin{tabular}{r|rrrr|rrrr}
 & \multicolumn{4}{c}{\textbf{Configuration}} & \multicolumn{4}{c}{\textbf{Metrics}}\\
\textbf{Id} & \textbf{B} & \textbf{M} & \textbf{R} & \textbf{H} & {\textbf{ARE}} & {\textbf{ATE}} & \textbf{IoU} & \textbf{Time}\\
& & & & & $10^{-3}$ $^{\circ}$/m & \% & \% & ms \\
\hline
1 & n & n & n & n & 50.5 			& 12.23 		& 84.88				& 69 \\
2 & y & n & n & n & 42.7 			& 9.98 			& 86.14				& 70 \\
3 & y & y & n & n & \textbf{33.7} 	& \textbf{9.73} & \textbf{87.80}	& 70 \\
4 & y & y & y & n & 49.8			& 12.48 		& 85.45				& \textbf{46} \\
5 & y & y & y & y & 47.9 			& 11.42 		& 85.69				& 55
\end{tabular}
\caption{
Evaluated network configurations and quantitative evaluation results based on the KITTI odometry and tracking benchmark.
We compare different techniques, changing one at a time: Gaussian blur (\textbf{B}), motion and spatial consistency loss (\textbf{M}), receptive field reduction (\textbf{R}) and half output resolution(\textbf{H}).
We evaluate the average rotation and translation error (\textbf{ARE} and \textbf{ATE}) based on the KITTI odometry benchmark.
We also determine the intersection-over-union (\textbf{IoU}) from the KITTI tracking benchmark by estimating vehicle positions in the second frame based on the ground truth positions in the first frame.
}
\label{tab:quantitative_results}
\end{table}

\paragraph{Training Procedure} \label{sec:application_to_grid_map_sequences_experiments_training_procedure}
We train all models on sequences 00-07 of the KITTI Odometry and sequences 00-08 of the KITTI tracking benchmark which resemble around 70\% of the overall data set.
For a better comparison, all models are trained for 52 epochs. 
With batch size 4 we train all configurations with a learning rate of $\mathrm{10^{-5}}$ for the first 21 epochs and then halving it after each 10 epochs.
To mitigate the imbalance due to mostly forward motion we also augment the input grid maps by applying random rotation in the range of [-10$^{\circ}$, 10$^{\circ}$] and random translation within [-3.5m, 3.5m].
The evaluation results are listed in Table~\ref{tab:quantitative_results}.

\paragraph{Results} \label{sec:application_to_grid_map_sequences_experiments_results}

\begin{figure}[ht]
\centering
\begin{subfigure}{\linewidth}
\centering
\fbox{\includegraphics[width=0.93\linewidth]{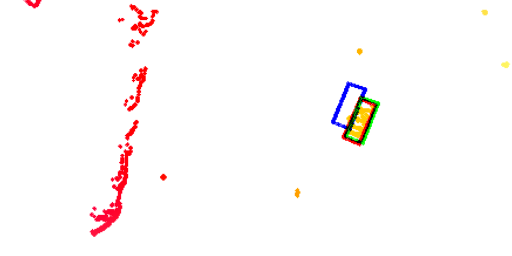}}
\end{subfigure}

\vspace{1mm}
\begin{subfigure}{\linewidth}
\centering
\fbox{\includegraphics[width=0.93\linewidth]{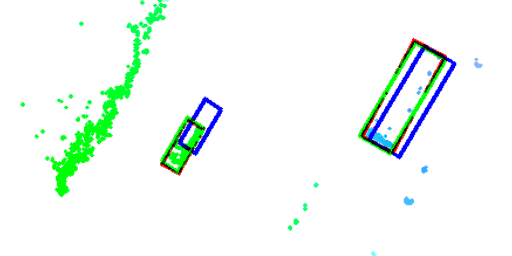}}
\end{subfigure}
\caption{
Qualitative result of the object prediction for the same scenario presented in Fig.~\ref{fig:qualitative_results} (top) and another scenario involving an oncoming car and a bus driving in the same direction (bottom).
Blue and red boxes denote the poses of labeled objects in the last and the current frame, respectively.
Green boxes depict objects of the last frame transformed by the average flow estimates of all object measurements of the last frame.
Grid maps artificially thickened for better visibility.
}
\label{fig:results_tracking}
\end{figure}

Figures~\ref{fig:results_odometry}~and~\ref{fig:results_tracking} depict qualitative results for odometry estimation and object prediction.
Applying Gaussian blur to Id~\textbf{2} improves the overall performance by reducing vanishing gradients during back-propagation.
Using the motion and spatial consistency losses for Id~\textbf{3} results in major improvement of performance at both estimating vehicle ego-motion and tracking performance.
Here, the average rotation error decreases by more than 20\%.
Especially for the tracking IoU we assume this might be due to a better separation between static and dynamic obstacles during flow estimation.
Although the improvement seems minor we note that there is a relatively small amount of moving traffic participants in the tracking data set.
Id~\textbf{4} with reduced receptive field size shows decreased performance.
As we reduce the receptive field size only depending on the vehicle ego-motion it might be too small to also cover the relative motion between oncoming vehicles.
However, it uses less memory and an inference time decreased by 35\%.
We also evaluated Id~\textbf{5} with one up-convolution layer to achieve a higher output scene flow resolution.
This configuration again decreases errors compared to the configuration with lower output resolution.

\section{Conclusion} \label{sec:conclusion}
We presented our approach to learn flow estimators based on deep convolutional networks in a self-supervised setting.
By extending the Unflow framework \cite{Meister2018} with domain knowledge on motion- and spatial consistency terms we were able to achieve better convergence properties and higher accuracy in the flow estimation task.
We showed that our method can be applied to camera image and grid map sequences.
In the future, we want to combine the learned feature representations of our self-supervised models with semantic tasks such as object detection where we aim to use only a few training examples.
For grid maps, another application would be the replacement of particle filter-based estimators by a learned scene flow estimator that encodes context information.

\bibliographystyle{IEEEtran}
\bibliography{root}

\end{document}